%% file: trf2017.tex
\documentclass{article}
\input{head}

\usepackage{spconf,amsmath,graphicx}

\title{Language modeling with Neural trans-dimensional random fields}
\name{Bin Wang, Zhijian Ou\thanks{This work is supported by NSFC grant 61473168.}}
\address{Department of Electronic Engineering, Tsinghua university, Beijing, China. \\
         wangbin12@mails.tsinghua.edu.cn, ozj@tsinghua.edu.cn}
\begin{document}
\ninept
\maketitle
\begin{abstract}
	
Trans-dimensional random field language models (TRF LMs) have recently been introduced, where sentences are modeled as a collection of random fields.
The TRF approach has been shown to have the advantages of being computationally more efficient in inference than LSTM LMs with close performance and being able to flexibly integrate rich features.
In this paper we propose neural TRFs, beyond of the previous discrete TRFs that only use linear potentials with discrete features. The idea is to use nonlinear potentials with continuous features, implemented by neural networks (NNs), in the TRF framework. Neural TRFs combine the advantages of both NNs and TRFs. The benefits of word embedding, nonlinear feature learning and larger context modeling are inherited from the use of NNs. At the same time, the strength of efficient inference by avoiding expensive softmax is preserved.
A number of technical contributions, including employing deep convolutional neural networks (CNNs) to define the potentials and incorporating the joint stochastic approximation (JSA) strategy in the training algorithm,
are developed in this work, which enable us to successfully train neural TRF LMs.
Various LMs are evaluated in terms of speech recognition WERs by rescoring the 1000-best lists of WSJ'92 test data.
The results show that neural TRF LMs not only improve over discrete TRF LMs, but also perform slightly better than LSTM LMs with only one fifth of parameters and 16x faster inference efficiency.

\end{abstract}
\section{Introduction}
\label{sec:intro}

Statistical language models, which estimate the joint probability of words in a sentence, form a crucial component in many applications such as automatic speech recognition (ASR) and machine translation (MT).
Recently, neural network language models (NN LMs), which can be either feedforward NNs (FNNs) \cite{schwenk2007continuous} or recurrent NNs (RNNs) \cite{mikolov2011,lstm2012}, have been shown to surpass classical n-gram LMs. RNNs with Long Short-Term Memory (LSTM) units are particularly popular.
Remarkably, both n-gram LMs and NN LMs follow the directed graphical modeling approach, which represents the joint probability in terms of conditionals.
In contrast, a new trans-dimensional random field (TRF) LM \cite{Bin2015,Bin2017} has recently been introduced in the undirected graphical modeling approach, where sentences are modeled as a collection of random fields and the joint probability is defined in terms of local potential functions.
It has been shown that TRF LMs significantly outperform n-gram LMs, and perform close to LSTM LMs but are computationally more efficient (200x faster) in inference (i.e. computing sentence probability).

Although the TRF approach has the capacity to support nonlinear potential functions and rich features, only linear potentials with discrete features (such as word and class n-gram features) are used in the previous TRF models, which limit their performances. The previous TRF models \cite{Bin2015,Bin2017} will thus be referred to as discrete TRFs.
This limitation is clear when comparing discrete TRF LMs with LSTM LMs.
First, LSTM LMs associate with each word in the vocabulary a real-valued feature vector.
Such word embedding in continuous vector space creates a notion of similarity between words and achieves a level of generalization that is hard with discrete features.
Discrete TRFs mainly rely on word classing and various orders of discrete features for smoothing parameter estimates.
Second, LSTM LMs learn nonlinear interactions between underlying features by use of NNs, while discrete TRF LMs basically are log-linear models.
Third, LSTM models could model larger contexts by using memory cells than discrete TRF models.
Despite these differences, discrete TRF LMs still achieves impressive performances, being close to LSTM LMs.
A promising extension is to integrate NNs into the TRF framework, thus eliminating the above limitation of discrete TRFs.

The above analysis motivates us to propose neural trans-dimensional random fields (neural TRFs) in this paper.
The idea is to use nonlinear potentials with continuous features, implemented by NNs, in the TRF framework.
Neural TRFs combine the advantages of both NNs and TRFs.
The benefits of word embedding, nonlinear feature learning and larger context modeling are inherited from the use of NNs.
At the same time, the strength of efficient inference by avoiding expensive softmax is preserved.

We have developed a stochastic approximation (SA) algorithm, called augmented SA (AugSA), with Markov chain Monte Carlo (MCMC) to estimate the model parameters and normalizing constants for discrete TRFs.
Note that the log-likelihood of a discrete TRF is concave, guaranteeing training convergence to the global maximum.
Fitting neural TRFs is a non-convex optimization problem, which is more challenging.
There are a number of technical contributions made in this work, which enable us to successfully train neural TRFs.
First, we employ deep convolutional neural networks (CNNs) to define the potential functions.
CNNs can be stacked to represent larger and larger context, and allows easier gradient propagation than LSTM RNNs.
Second, the AugSA training algorithm is extended to train neural TRFs, by incorporating the joint stochastic approximation (JSA) \cite{xu2016joint} strategy, which has been used to successfully train deep generative models.
The JSA strategy is to introduce an auxiliary distribution to serve as the proposal for constructing MCMC operator for the target distribution.
The log-likelihood of the target distribution and the KL-divergence between the target distribution and the auxiliary distribution are jointly optimized.
The resulting AugSA plus JSA algorithm is crucial for handling deep CNN features, not only significantly reducing computation cost for every SA iteration step but also
considerably improving SA training convergence.
Third, several additional techniques are found to improve the convergence for training neural TRFs, including wider local jump in MCMC, Adam optimizer \cite{adam}, and training set mini-batching.

Various LMs are evaluated in terms of speech recognition WERs by rescoring the 1000-best lists of WSJ'92 test data.
The neural TRF LM improves over the discrete TRF LM, reducing WER from 7.92\% to 7.60\%, with less parameters.
Compared with state-of-the-art LSTM LMs \cite{lstmdropout}, the neural TRF LM outperforms the small LSTM LM (2 hidden layers and 200 units per layer) with relative WER reduction of 4.5\%,
and performs slightly better than the medium LSTM LM (2 hidden layers and 650 units per layer) with only one fifth of parameters.
Moreover, the inference of the neural TRF LM is about 16 times faster than the medium LSTM LM. The average time cost for rescoring a 1000-best list for a utterance in WSJ'92 test set are about 0.4 second vs 6.36 seconds, both using 1 GPU.

In the rest of the paper, we first discuss related works in Section \ref{sec:RelatedWork}. Then we introduce the new neural TRF model in Section \ref{sec:ModelDefination} and its training algorithm in Section \ref{sec:ModelLearning}. After presenting  experimental results in Section \ref{sec:Experiments}, the conclusions are made in Section \ref{sec:conclusion}.

\section{Related work}
\label{sec:RelatedWork}

LM research can be roughly divided into two tracks. The directed graphical modeling approach, includes the classic n-gram LMs and various NN LMs. The undirected graphical modeling approach, has few priori work, except \cite{rosenfeld2001whole,Bin2015,Bin2017}.
A review of the two tracks can be found in \cite{Bin2017}.
To our knowledge, the TRF work represents the first success in using undirected graphical modeling approach to language modeling.
Starting from discrete TRFs, the main new features of neural TRFs proposed in this paper is the marriage of random fields and neural networks, and the use of CNNs for feature extraction.
In the following, we mainly comment on these two related studies and the connection to our work.

\subsection{Marriage of random fields and neural networks}

A key strength of NNs is their nonlinear feature learning ability. Random fields (RFs) are powerful for describing interactions among structured random variables.
Combining RFs and NNs has been pursued but
most models developed so far are in fact combining conditional random fields (CRFs) \cite{CRF2001} and NNs, namely ``CRFs+NNs''.
In conventional CRFs, both node potentials and edge potentials are defined as linear functions using discrete indicator features.
``CRFs+NNs'' has been introduced a few times in priori literature.
It was termed Conditional Neural Fields in \cite{peng2009conditional}, and later Neural Conditional Random Fields \cite{artieres2010neural} with a slightly different specification of potentials.
It also appeared previously in the speech recognition
literature \cite{prabhavalkar2010backpropagation}.
Recently, there are increasing more studies that use various types of NNs, e.g. FNNs \cite{collobert2011natural}, RNNs \cite{yao2014recurrent}, LSTM RNNs \cite{ma2016end}, to extract features as input to CRFs.
Remarkably, the general idea in ``CRFs+NNs'' models is to implement the combination by using NNs to represent the potential functions in a RF.
This is in spirit the same as in neural TRFs.
However the algorithms developed in ``CRFs+NNs'' studies are not applicable to neural TRFs because the sample spaces of these CRFs are much smaller than that of TRFs.

It is worth pointing out that CRFs can only be used for discriminative tasks, e.g. sequence labeling, structured prediction tasks. In contrast, language modeling is a generative modeling task.
A generative random field model is proposed in \cite{xie2016theory}, where the potentials are also defined as CNNs, but it is only for modeling fixed-size images (i.e. fix-dimensional modeling).

\subsection{Convolutional neural networks}

Besides the great success in computer vision, CNNs have recently received more attention in language modeling.
CNNs over language act as feature detectors and can be stacked hierarchically to capture large context, like in computer vision.
It is shown in \cite{pham2016convolutional} that applying convolutional layers in FNNs performs better than conventional FNN LMs but is below LSTM LMs.
Convolutional layers can also be used within RNNs, e.g. as studied in \cite{kim2016character}, 1-D convolutional filters of varying widths are applied over characters, whose output is fed to the upper LSTM RNN.
Recently, it is shown in \cite{dauphin2016language} that CNN-FNNs with a novel gating mechanism benefit gradient propagation and perform slightly better than LSTM LMs. Similarly, our pilot experiment shows that using the stacked structure of CNNs in neural TRFs allows easier model training than using the recurrent structure of RNNs.

\section{Model Definition}
\label{sec:ModelDefination}

Throughout, we denote by $x^l=(x_1, \cdots, x_l)$ a sentence (i.e. word sequence) of length $l$, ranging from $1$ to $m$.
Sentences of length $l$ are assumed to be distributed from an exponential family model:
    \begin{equation} \label{eq:model_l}
    p_l(x^l;\theta) = \frac{1}{Z_l(\theta)} e^{\phi(x^l;\theta)}
    \end{equation}
where $\theta$ indicates the set of parameters and
$\phi$ is the potential function,
and $Z_l(\theta)$ is the normalization constant of length $l$, i.e. $Z_l(\theta) = \sum_{x^l} e^{\phi(x^l; \theta)}$.
Moreover, assume that length $l$ is associated with a probability $\pi_l$ for $l=1, \cdots, m$. Therefore, the pair $(l,x^l)$ is jointly distributed as:
    \begin{equation} \label{eq:model_all}
    p(l, x^l;\theta) = \pi_l p_l(x^l; \theta)
    \end{equation}

\begin{figure}[t]
	\centering
	\includegraphics[width=\linewidth]{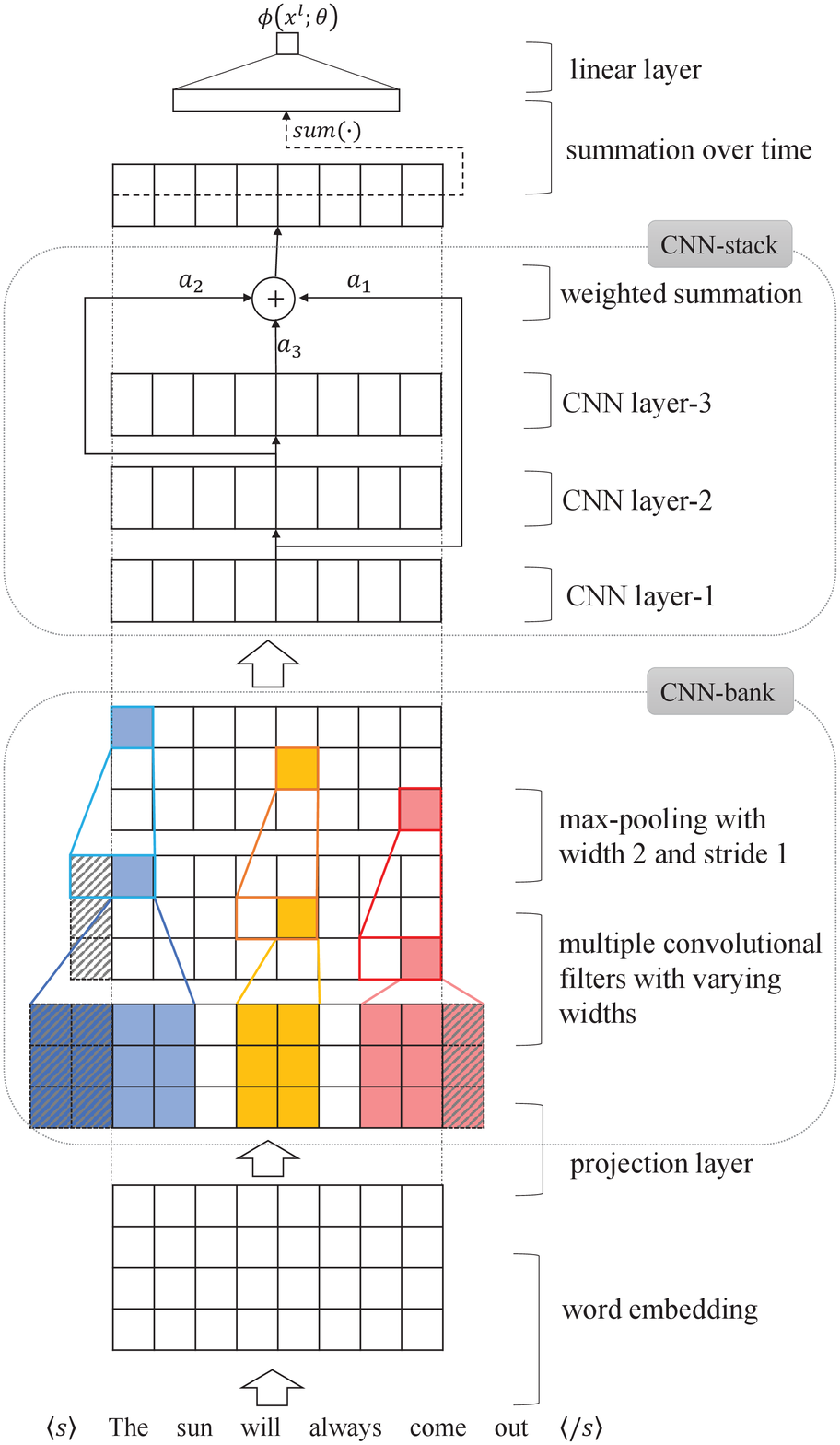}
	\caption{The deep CNN architecture used to define the potential function $\phi(x^l;\theta)$.
		Shadow areas denote the padded zeros.
	}
	\label{fig:cnn}
	\vspace{-10pt}
\end{figure}

Different from using linear potentials in discrete TRFs \cite{Bin2017}, neural TRFs define the potential function $\phi(x^l;\theta)$ by a deep CNN, as described below and shown in Fig. \ref{fig:cnn}.

\textbf{Embedding and projection.}
First, each word $x_i$ ($i=1, \cdots, l$) in a sentence is mapped to an embedded vector $e_i \in R^{d_e}$.
Then a projection layer with rectified linear unit (ReLU) activation is applied to each embedding vector to reduce the dimension, i.e.
    \begin{equation}
    y_i = \max\{W_{p} e_i + b_{p}, 0\}, i=1, \cdots, l
    \end{equation}
where $y_i \in R^{d_p}$ is the output of the projection layer of dimension $d_p$;
$W_{p} \in R^{d_p \times d_e}$ and $b_{p} \in R^{d_p}$ are the parameters.

\textbf{CNN-bank.}
The outputs of the projection layer are fed into a CNN-bank module, which contains a set of 1-D convolutional filters with widths ranging from 1 to $K$.
These filters explicitly model local contextual information (akin to modeling unigrams, bigrams, up to $K$-grams) \cite{kim2016character}.
Denote by $F \in R^{d_p \times k}$ a filter of width $k$, $k=1, \cdots, K$, and
by $Y = [y_1, \cdots , y_l] \in R^{d_p \times l}$ the output of the projection layer.
We first symmetrically pad zeros to the beginning and end of $Y$ to make it $k-1$ longer, denoted by $Y' \in R^{d_p \times (l+k-1)}$.
Then the convolution is performed between $Y'$ and filter $F$, and the output \emph{feature map} $f \in R^l$ is given by
    \begin{equation}
    f[i] = \max\{\langle Y'[:, i:i+k-1], F \rangle, 0\}, i=1, \cdots, l
    \end{equation}
where $f[i]$ is the $i$-th component of vector $f$,
$Y'[:, i:i+k-1]$ is the $i$-to-$(i+k-1)$-th columns of $Y'$ and
$\langle A, B \rangle$ is the Forbenius inner product.
Such convolution with the above padding scheme is known as \emph{half convolution}\footnote{http://deeplearning.net/software/theano/library/tensor/nnet/conv.html}.
The output feature maps from multiple filters with varying widths are spliced together,
and followed by a max-pooling over time with width 2 and stride 1.
Zeros are also padded before the max-pooling to preserve the time dimensionality.
Suppose there are $w$ filters for each filter width.
The output of the CNN-bank module is $Y_{b} \in R^{wK \times l}$.

\textbf{CNN-stack.}
On the top of the CNN-bank module, a CNN-stack module consists of a stack of 1-D convolutional layers to further extract hierarchical features with $n$ layers.
The outputs of each convolutional layer are weighted summarized, which is similar to the skip-connections in \cite{van2016wavenet}.
Let $Y_s^j \in R^{d_{s} \times l}$ denote the output of the $j$-th convolutional layer, $j=1, \cdots, n$.
Our experiments use $d_s < wK$ to reduce the dimension and employ \emph{half convolution} at each CNN layer.
The output of the CNN-stack module $Y_s \in R^{d_s \times l}$ is:
    \begin{equation}
    Y_{s}[:,i] = \max\{0, \sum_{j=1}^n a_j \ast Y_s^j[:,i]\}, i = 1, \cdots, l
    \end{equation}
where $a_j \in R^{d_{s}}$ is the weights applied to the $j$-th layer and $\ast$ is the element-wise multiplication.

\textbf{Summation over time.}
Finally, the potential function is defined to take the following value:
    \begin{equation}
    \phi(x^l; \theta) = \lambda^T \sum_{i=1}^l Y_{s}[:,i] + c
    \end{equation}
where $\lambda \in R^{d_s}$, $c \in R$ are the parameters.
In summary, $\theta$ denotes the collection of all the parameters defined in the deep CNNs.

\section{Model learning}
\label{sec:ModelLearning}

A novel learning algorithm, called augmented SA (AugSA), has been developed to estimate both the model parameters and normalizing constants for discrete TRFs \cite{Bin2017}.
In this section, the AugSA algorithm is extended to train neural TRFs.

\subsection{AugSA plus JSA}

In AugSA, we introduce the following joint distribution of $(l,x^l)$:
    \begin{equation} \label{eq:model_all_zeta}
    p(l, x^l;\theta, \zeta) = \pi_l^0 p_l(x^l;\theta,\zeta) = \frac{\pi_l^0}{Z_1(\theta) e^{\zeta_l}} e^{\phi(x^l;\theta)}
    \end{equation}
where $\zeta=(\zeta_1, \cdots, \zeta_m)^T$ with $\zeta_1=0$ and $\zeta_l$ is the hypothesized value of the log ratio of $Z_l(\theta)$ with respect to $Z_1(\theta)$, namely $\log\{ Z_l(\theta)/Z_1(\theta)\}$, $l=1,\cdots,m$.
$Z_1(\theta)$ is chosen as the reference value and can be calculated exactly.
$\pi_l^0$ is the specified length probability used in model training.
Note that we set the prior length probability $\pi_l$ to the empirical length probability in inference.

Denote by $D$ the training set and by $D_l$ the collection of sentences of length $l$ in the training set. The maximum likelihood estimation of parameter $\theta$ and normalization constant $\zeta$ can be found by solving the following simultaneous equations \cite{Bin2017}:
    \begin{align}
        &E_{D} \left[ \pd{\phi}{\theta} \right] -
        \sum_{l=1}^{m} \frac{|D_l|}{|D|} E_{p_l}\left[ \pd{\phi}{\theta} \right]
        = 0, \label{eq:obj1}  \\
        &\sum_{x^l} p(l,x^l;\theta,\zeta) = \pi^0_l, \label{eq:obj2}
    \end{align}
where $|D_l|$ is the number of sentences in set $D_l$, $|D| = \sum_{l=1}^{m} |D_l|$,
$E_D$ is the empirical expectation on the training set $D$
and $E_{p_l}$ is the expectation with respect to the model distribution $p_l(x^l;\theta,\zeta)$ in \eqref{eq:model_all_zeta}:
$$E_{D} \left[ \pd{\phi}{\theta} \right] = \frac{1}{|D|} \sum_{l=1}^{m} \sum_{x^l \in D_l} \pd{\phi(x^l;\theta)}{\theta}$$
$$E_{p_l}\left[ \pd{\phi}{\theta} \right]= \sum_{x^l} p_l(x^l;\theta,\zeta) \pd{\phi(x^l;\theta)}{\theta}$$


Exact solving \eqref{eq:obj1} and \eqref{eq:obj2} is infeasible.
AugSA is proposed to stochastically solve \eqref{eq:obj1} and \eqref{eq:obj2} in the SA framework \cite{SA51}, which iterates MCMC sampling and parameter update.
The convergence of SA has been studied under various conditions \cite{benveniste2012adaptive,chen2002stochastic,Tan2015}.
The MCMC sampling in AugSA is implemented by the trans-dimensional mixture sampling (TransMS) algorithm \cite{Bin2017} to simulate sentences of different dimensions from the joint distribution $p(l,x^l;\theta,\zeta)$ in \eqref{eq:model_all_zeta}.

The sampling operations in TransMS make the computational bottleneck in AugSA.
Suppose that we use Gibbs sampling to simulate a sentence for a given length, which is the method used in \cite{Bin2017} for training discrete TRFs.
We need to calculate the conditional distribution $p_l(x_i|x_{\neq i})$ of word $x_i$ for each position $i$, given all the other words $x_{\neq i}$.
This is computational expensive because calculating $p_l(x_i|x_{\neq i})$ needs to enumerate all the possible values of $x_i \in \mathcal{V}$ and to compute the joint probability $p_l(x_i, x_{\neq i})$ for each possible value, where $\mathcal{V}$ denotes the vocabulary.
In \cite{Bin2017}, word classing is introduced to accelerate sampling, which means that each word is assigned to a single class.
Through applying Metropolis-Hastings (MH) within Gibbs sampling, we first sample the class by using a reduced model as the proposal, which includes only the features that depend on $x_i$ through its class, and then sample the word. This reduces the computational cost from $|\mathcal{V}|$ to $|\mathcal{V}|/|\mathcal{C}|$, where $|\mathcal{C}|$ denotes the number of classes.
However, the computation reduction in using word classing in neural TRFs is not as significant as in discrete TRFs, because the deep CNN potentials in neural TRFs involve a much larger context, which makes the sampling computation with the reduced model still expensive.

\begin{figure}
\begin{algorithmic}[1]
\Require training set $D$
\State Init the parameter $\theta^{(0)}$ and $\mu^{(0)}$ and set

    $\zeta^{(0)} = (0, \log{|\mathcal{V}|}, 2\log{|\mathcal{V}|}, \cdots, (m-1)\log{|\mathcal{V}|})$,
    where $|\mathcal{V}|$ is the vocabulary size.

\For{$t=1,2,\dots,t_{max}$}
    \State Random select $K_D$ sentences from the training set, as $D^{(t)}$
    \State Generate $K_B$ sentences using TransMS in section \ref{sec:sampling}, as $B^{(t)}$

    \State Compute $\theta^{(t)}$ based on \equref{eq:uplambda}   \label{al:jsa:uplambda}
    \State Compute $\zeta^{(t)}$ based on \equref{eq:upzeta1} and \equref{eq:upzeta2} \label{al:jsa:upzeta}
    \State Compute $\mu^{(t)}$ based on \equref{eq:upmu}
\EndFor

\end{algorithmic}
\caption{The AugSA plus JSA algorithm for training neural TRFs} \label{fig:AugSA}
\end{figure}

To apply AugSA to neural TRFs, we borrow the idea of joint stochastic approximation (JSA) \cite{xu2016joint}, which has been used to successfully train deep generative models.
The JSA strategy is to introduce an auxiliary distribution $q(l, x^l; \mu)$ with parameter $\mu$ to serve as the proposal for constructing MCMC operator for the target distribution $p(l,x^l;\theta,\zeta)$ in \eqref{eq:model_all_zeta}.
The log-likelihood of the target distribution and the KL-divergence between the target distribution and the auxiliary distribution are jointly optimized.
Therefore, the AugSA plus JSA algorithm is defined as stochastically solving the three simultaneous equations, namely \eqref{eq:obj1}, \eqref{eq:obj2} together with
    \begin{equation} \label{eq:objKL}
    \pdone{\mu} KL(p(l, x^l; \theta, \zeta) || q(l, x^l; \mu)) = 0
    \end{equation}
At each iteration, the parameter $\mu$ of the auxiliary distribution $q(l,x^l;\mu)$ is updated together with the parameter $\theta$ and normalization constants $\zeta$,
and $q(l,x^l;\mu)$ is used in TransMS as a proposal distribution (See Section \ref{sec:sampling} for details).
In this paper, the auxiliary distribution is implemented by an LSTM RNN.

Moreover, several additional techniques are used to suit the use of nonlinear potentials and to improve the convergence for training neural TRFs.
The first technique, called training set mini-batching \cite{Bin2015}, is that at each iteration, a mini-batch of sentences is randomly selected from the training set and the empirical expectation is calculated over this mini-batch.
This is crucial for training neural TRFs because unlike in discrete TRFs, the gradient of the nonlinear potential function $\phi$ with respect to $\theta$ depends on the parameters $\theta$.
Second, the Adam \cite{adam} optimizer is used to update the parameter $\theta$, saving the computation cost for estimating the empirical variances.

The AugSA plus JSA algorithm is summarized in Fig. \ref{fig:AugSA} and detailed as follows.
At iteration $t$, we first random select $K_D$ samples from training set $D$, denoted by $D^{(t)}$.
Then TransMS (in Section \ref{sec:sampling}) is performed to generate $K_B$ samples, denoted by $B^{(t)}$.
The update for the parameters $\theta$ is:
    \begin{equation} \label{eq:uplambda}
    \theta^{(t)} = \theta^{(t-1)} + \gamma_{\theta,t} Adam \left\{
        E_{D^{(t)}} \left[ \pd{\phi}{\theta} \right] -
        \frac{1}{K_B}\sum_{(l,x^l) \in B^{(t)}} \frac{\tilde{\pi}_l}{\pi_l^0} \pd{\phi}{\theta} \right\},
    \end{equation}
where $Adam$ denotes the Adam optimizer, $\gamma_{\theta,t}$ is the learning rate for $\theta$ and $\tilde{\pi}_l$ is the empirical probability of length $l$.
Remarkably, the gradient of $\phi$ with respect to $\theta$ can be efficiently computed by backpropagtion.
The update for the normalization constants $\zeta$ is the same as in \cite{Bin2017}:
    \begin{align}
     \zeta^{(t-\frac{1}{2})} & = \zeta^{(t-1)} + \gamma_{\zeta,t} \left\{ \frac{\delta_1(B^{(t)})}{\pi^0_1}, \cdots, \frac{\delta_m(B^{(t)})}{\pi^0_m} \right\}^T, \label{eq:upzeta1} \\
     \zeta^{(t)} & = \zeta^{(t-\frac{1}{2})} - \zeta^{(t-\frac{1}{2})}_1, \label{eq:upzeta2}
    \end{align}
where $\zeta^{(t)}_1$, the first element of $\zeta^{(t)}$, is set to 0 by \eqref{eq:upzeta2},
and $\gamma_{\zeta,t}$ is the learning rate for $\zeta$,
and $\delta_l(B^{(t)})$ is the proportion of length $l$ appearing in $B^{(t)}$:
    \begin{equation*}
    \delta_l(B^{(t)}) = \frac{1 }{K_B} \sum_{(j,x^j) \in B^{(t)}} { 1(j=l) } .
    \end{equation*}
The update for the parameters $\mu$ of the auxiliary distribution is:
    \begin{equation} \label{eq:upmu}
    \mu^{(t)} = \mu^{(t-1)} + \gamma_{\mu,t} \sum_{(l,x^l) \in B^{(t)}} \pdone{\mu} \log q(l,x^l;\mu),
    \end{equation}
where $\gamma_{\mu,t}$ is the learning rate for $\mu$.

\subsection{TransMS with an auxiliary distribution}
\label{sec:sampling}

The trans-dimensional mixture sampling (TransMS) proposed in \cite{Bin2017} is extended for applying AugSA plus JSA.
The TransMS consists of two steps at each iteration: local jump between dimensions and Markov move for a given dimension.
First, the auxiliary distribution $q(l, x^l; \mu)$ is introduced as the proposal distribution $g(\cdot|x^h)$ in both local jump and Markov move.
Second, multiple-trial metropolis independence sampling (MTMIS) \cite{liu2008} is used to increase the acceptance rate.
Let $l^{(t-1)}$ and $x^{(t-1)}$ denote the length and sequence before sampling at iteration $t$.
The algorithm is described as follows.


\textbf{Step I: Local jump.}
Assuming $l^{(t-1)}=k$, first we draw a new length $j \sim \Gamma(k,\cdot)$,
where the jump distribution $\Gamma(k,\cdot)$ is defined to be uniform in the neighborhood of $k$:
    \begin{equation}
    \Gamma(k,j) =
        \frac{1}{\min(k+r, m) - \max(k-r,1) + 1},
    \end{equation}
if $|k-j| \leq r$ and to be $0$ otherwise,
where $m$ is the maximum length and $r$ is the jump range, which is set to be one in \cite{Bin2017}.

If $j=k$, we retain the observation, i.e. set $l^{(t)} = l^{(t-1)}$ and $x^{(t)} = x^{(t-1)}$, and perform the next step directly.

If $j>k$, we first draw a subsequence $u^{j-k}$ of length $j-k$ from a proposal distribution:
$u^{j-k} \sim g(\cdot|x^{(t-1)})$.
Then we set $l^{(t)} = j$ and $x^{(t)} = \{x^{(t-1)}, u^{j-k}\}$ with probability
    \begin{equation}\label{eq:ac1}
    \min \left\{ 1,
        \frac{\Gamma(j,k)}{\Gamma(k,j)}
        \frac{p(j, \{x^{(t-1)}, u^{j-k}\};\theta,\zeta)}
             {p(k, x^{(t-1)};\theta,\zeta) g(u^{j-k}|x^{(t-1)})}
    \right\},
    \end{equation}
where $\{x^{(t-1)}, u^{j-k}\}$ denotes the sequence with length $j$ whose first $k$ elements are $x^{(t-1)}$ and the last $j-k$ elements are $u^{j-k}$.

If $j<k$, we set $l^{(t)} = j$ and $x^{(t)} = x^{(t-1)}_{1:j}$ with probability
    \begin{equation}\label{eq:ac2}
    \min \left\{ 1,
        \frac{\Gamma(j,k)}{\Gamma(k,j)}
        \frac{p(j,x_{1:j}^{(t-1)};\theta,\zeta) g(x_{j+1:k}^{(t-1)}|x_{1:j}^{(t-1)})}
             {p(k,x^{(t-1)};\theta,\zeta)}
        \right\},
    \end{equation}
where $x_{i:j}^{(t-1)}$ denotes the subsequence of $x^{(t-1)}$ from the $i$-th position to the $j$-th position.

Here, the proposal distribution $g(\cdot|x^h)$ is computed using the auxiliary distribution $q(l,x^l;\mu)$.
For $q(l,x^l;\mu)$ implemented by an LSTM RNN as in this paper, calculating $g(\cdot|x^h)$ and sampling form $g(\cdot|x^h)$ can be performed recurrently.

\textbf{Step II: Markov move.}
In this step, given the current sequence and fixing the length, we perform the block Gibbs sampling from the first position to the last position with block-size $s$.
MTMIS is used with the same proposal distribution $g(\cdot|x^h)$ as in local jump. In our experiments,  the block-size $s=5$ and multiple-trial number $M=10$.
Denote by $(l, x^l)$ the current length and sequence after local jump. For positions $i=1, s+1, 2s+1, \cdots$, Markov move proceeds as follows:
\begin{itemize}
  \item Generate $M$ i.i.d. samples $u^s_{(k)} \sim g(\cdot|x^l_{1:i-1})$, each of length $s$, $k=1,\cdots,M$, compute
        $$w(u^s_{(k)})=\frac{p(l, \{x^l_{1:i-1}, u^s_{(k)}, x^l_{i+s:l}\}; \theta, \zeta)}
                            {g(u^s_{(k)}|x^l_{1:i-1})}
        $$
        and $W = \sum_{k=1}^{M} w(u^s_{(k)})$,
  where $u^s_{(k)}$ is a sequence of length $s$ and $\{x^l_{1:i-1}, u^s_{(k)}, x^l_{i+s:l}\}$ denotes the concatenation of the three subsequences $x^l_{1:i-1}$, $u^s_{(k)}$ and $x^l_{i+s:l}$.
  \item Draw $u^s$ from the trial set $\{u^s_{(1)}, \cdots, u^s_{(M)}\}$ with probability proportional to $w(u^s_{(k)})$.
  \item Set $x^{(t+1)} = \{x^l_{1:i-1}, u^s, x^l_{i+s:l}\}$ with probability
        $$
            \min\left\{1, \frac{W}{W - w(u^s) + w(x^l_{i: i+s-1})} \right\}
        $$
\end{itemize}

\section{Experiments}
\label{sec:Experiments}

\subsection{Experiment setup and baseline LMs}
\label{sec:ptb}

\begin{table}
    \centering
    \begin{tabular}{c|c|c|c|c}
    \hline
    LSTMs & PPL & WER-p & WER-r & WER-s \\
    \hline \hline
    LSTM-2$\times$200   &   113.9   &   8.14    &   7.96    &   7.80 \\
    \hline
    LSTM-2$\times$650   &   84.1    &   7.59    &   7.66    &   7.85 \\
    \hline
    LSTM-2$\times$1500  &   78.7    &   7.32    &   7.36    &   8.18 \\
    \hline
\end{tabular}
    \caption{Comparison of three different rescoring strategies for LSTM LMs.
    ``LSTM-2$\times$200/650/1500'' indicates the LSTM with 2 hidden layers and 200/650/1500 hidden units per layer.
    ``PPL'' denotes the perplexity on the PTB test set.
    ``WER-p'', ``WER-r'' and ``WER-s'' correspond to the preserve, reset and shuffle strategy respectively.
    }
    \label{tab:lstmwer}
\end{table}

In this section, we compare neural TRF LMs with different LMs on speech recognition.
The LM training corpus is the Wall Street Journal (WSJ) portion of Penn Treebank (PTB).
Sections 0-20 are used as the training set (about 930 K words),
sections 21-22 as the development set (74 K) and section 23-24 as the test set (82 K).
The vocabulary is limited to 10 K words, including a special token $\langle\text{UNK}\rangle$ denoting the word not in the vocabulary.
This setting is the same as that used in other studies \cite{mikolov2011, lstmdropout, Bin2015, Bin2017}.
For evaluation in terms of speech recognition WERs,
various LMs obtained using PTB training and development sets are applied to rescore the 1000-best lists from recognizing WSJ'92 test data (330 utterances).
For each utterance, the 1000-best list of candidate sentences are generated by the first-pass recognition using the Kaldi toolkit \cite{Kaldi} with a DNN-based acoustic models.
The oracle WER of the 1000-best list is 0.93\%.

The baseline LMs include a 5-gram LM with modified Kneser-Ney smoothing \cite{chen1999empirical} (denoted by ``KN5''),
and three LSTM LMs with 2 hidden layers and 200, 600, 1500 hidden units per layer respectively, which are called small, medium and large LSTMs in \cite{lstmdropout}.
We reproduce the three LSTM LMs in \cite{lstmdropout} and use them in our rescoring experiments.
The 5-gram LM is trained using the SRILM toolkit \cite{SRILM}, which automatically adds the beginning-token $\langle\text{s}\rangle$ and end-token $\langle\text{/s}\rangle$ at the beginning and the end of each sentence.
When applying the KN5 LM to rescoring, the beginning-token and end-token are also added to each sentence in the 1000-best list.

In contrast, for training LSTM LMs, a tricky practice \cite{lstmdropout} is to only add the end-token at the end of each sentence, and then concatenate all training sentences.
This in fact treats the whole training corpus as a single long sentence.
In rescoring, only the end-token is added to each sentence, like in training. But there are three strategies of how the initial hidden state is configured in rescoring.
\begin{enumerate}
  \item preserve: the final hidden state (i.e. the hidden state which predicts the end-token) of the previous sentence is preserved and used to compute the initial state of the current sentence together with the end-token.
  \item reset: the final state of the previous sentence is reset to zero.
  \item shuffle: we first shuffle all the candidate sentences of all the testing utterances, and then use the first strategy. This
  eliminate the unfair use of information across sentences.
\end{enumerate}

The WERs of the above strategies are shown in Table \ref{tab:lstmwer}.
It can be seen that the lowest WER is achieved by ``LSTM-2$\times$1500'' with 2 hidden layer and 1500 hidden units, using the preserve strategy.
When we shuffle the sentences, the WER increases significantly, from 7.32 to 8.18, which is even worse than the WER of ``LSTM-2$\times$200''.
This is because that in the preserve strategy, the candidate sentences of each utterance are rescored successively. The final hidden state carries relevant information to benefit the prediction of the next candidate sentence which belongs to the same testing utterance as the current candidate sentence.
After shuffling, the relation between adjacent candidate sentences is broken. The information preserved in the hidden states may mislead the prediction.

In the following, we use ``WER-r'', the WER obtained by the reset strategy, as the performance measure of the LSTM LMs, since the resulting WERs are independent of the processing order of the candidate sentences and stable.
Moreover, this enables a more fair comparison with KN5 and neural TRF LMs, since they do not use any information across sentences.

\begin{table*}
	\centering
	\begin{tabular}{l|c|c|c|l|l}
		\hline
		Model               &   PPL     &   WER(\%)     &   \#param (M)  & Training time  & Inference time \\
		\hline
		KN5                 &   141.2  &   8.78        &   2.3        & 22 seconds (1 CPU)  &  0.06 seconds (1 CPU)\\
		LSTM-2$\times$200   &   113.9   &   7.96    &   4.6          & about 1.7 hours (1 GPU)   &  6.36 seconds (1 GPU)\\
		LSTM-2$\times$650   &   84.1    &   7.66    &   19.8        & about 7.5 hours (1 GPU)   &  6.36 seconds (1 GPU)\\
		LSTM-2$\times$1500  &   78.7    &   7.36    &   66.0        & about 1 day (1 GPU)       &  9.09 seconds (1 GPU)\\
		discrete TRF \cite{Bin2017} & $\geq$130    &   7.92    &   6.4        & about 1 day (8 CPUs)      &  0.16 seconds (1 CPU)\\
		\hline
		neural TRF             &   $\geq$37.4  &   7.60   &   4.0      & about 3 days (1 GPU)       &  0.40 second (1 GPU)\\
		\hline
		KN5$+$LSTM-2$\times$1500  &   -    &   7.47   &   & \\
		TRF$+$LSTM-2$\times$1500  &   -    &   7.17   &   & \\
		\hline
	\end{tabular}
	\caption{Performances of various LMs.
		``PPL'' is the perplexity on PTB test set.
		``WER'' is the word error rate on WSJ'92 test data.
		``\#param'' is the number of parameter numbers (in millions).
		``$+$'' denotes the log-linear interpolation with equal weights of $0.5$.
		For LSTMs, the WER is obtained using the reset strategy.
		``Inference time'' denotes the average time of rescoring the 1000-best list for each utterance.
	}
	\label{tab:wer}
\end{table*}

\subsection{Neural TRF LMs in speech recognition}

\begin{table}
    \centering
    \begin{tabular}{c|l}
        \hline
        word embedding size  & 256 \\
        \hline
        projection dimension & 128 \\
        \hline
        CNN-bank   & cnn-$k$-128, with $k$ ranging from 1 to 10 \\
        \hline
        max-pooling & width 2 and stride 1 \\
        \hline
        CNN-stack & cnn-$3$-128 $\rightarrow$ cnn-$3$-128 $\rightarrow$ cnn-$3$-128 \\
        \hline
    \end{tabular}
    \caption{CNN configuration in neural TRFs.
    ``cnn-$k$-$n$'' denotes a 1-D CNN with filter width $k$ and output dimension $n$.
    ``$A \rightarrow B$'' denotes that the output of layer $A$ is fed into layer $B$.
    }
    \label{tab:cnnconfig}
\end{table}

The CNN configuration used in neural TRF models is shown in table \ref{tab:cnnconfig}.
The AugSA plus JSA algorithm in Fig.\ref{fig:AugSA} is used to train neural TRF LMs on PTB training corpus.
At each iteration, we random select $K_D = 1000$ sentences from training corpus,
and generate $K_B=100$ sentences of various lengths using the TransMS algorithm described in Section \ref{sec:sampling}.
The auxiliary distribution $q(l, x^l;\mu)$ is defined as a LSTM LM with 1 hidden layers and 250 hidden units.
The learning rates in \eqref{eq:uplambda}, \eqref{eq:upzeta1} and \eqref{eq:upmu} are set as
$\gamma_{\theta,t} = 1/(t + 10^4)$,
$\gamma_{\zeta, t} = t^{-0.2}$ and $\gamma_{\mu, t} = 1.0$.
The length distribution $\pi^0_l$ is set as specified in \cite{Bin2017}.

\begin{figure}[t]
\begin{minipage}[b]{.48\linewidth}
  \centering
  \centerline{\includegraphics[width=\linewidth]{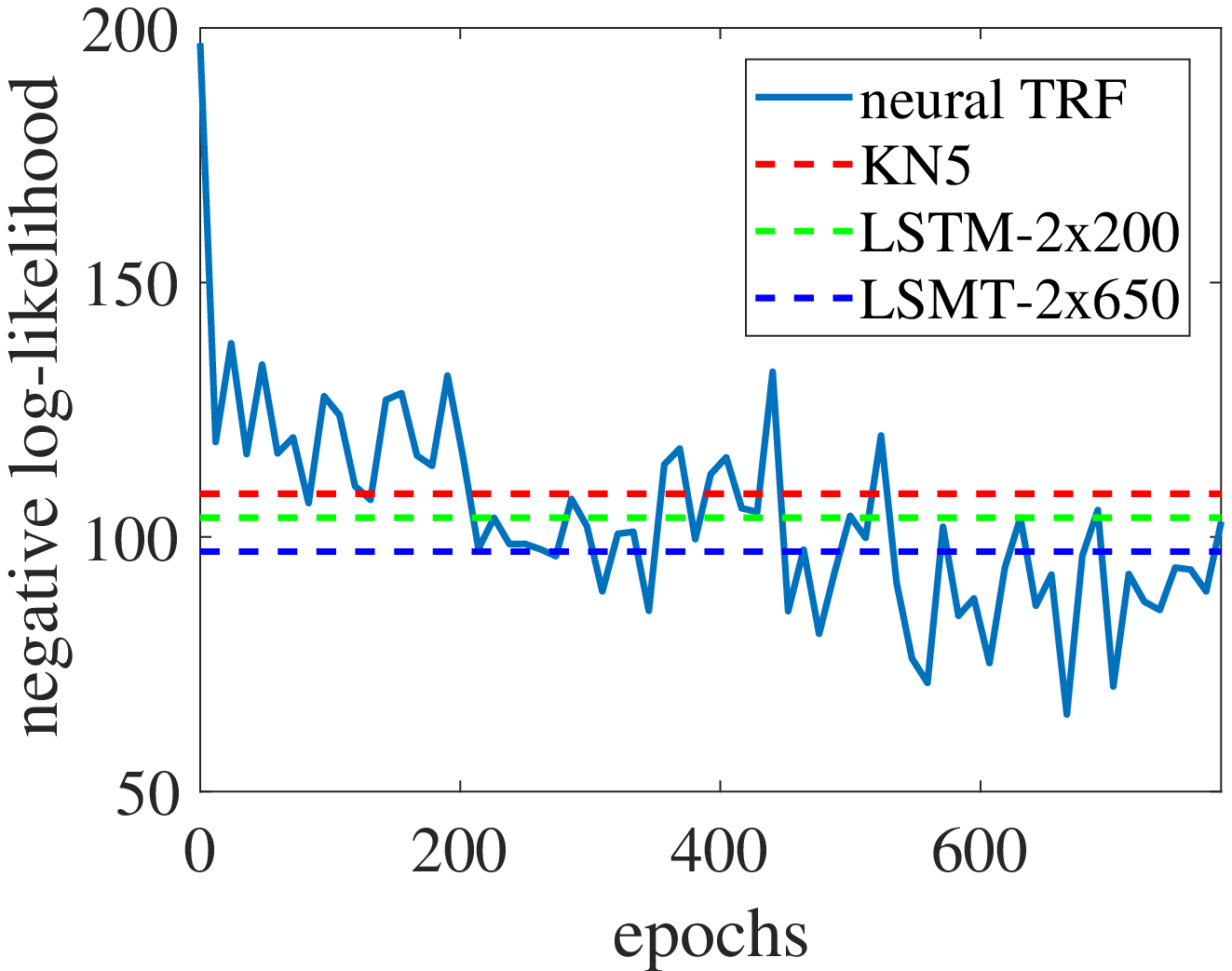}}
  \centerline{(a)}
\end{minipage}
\hfill
\begin{minipage}[b]{0.48\linewidth}
  \centering
  \centerline{\includegraphics[width=\linewidth]{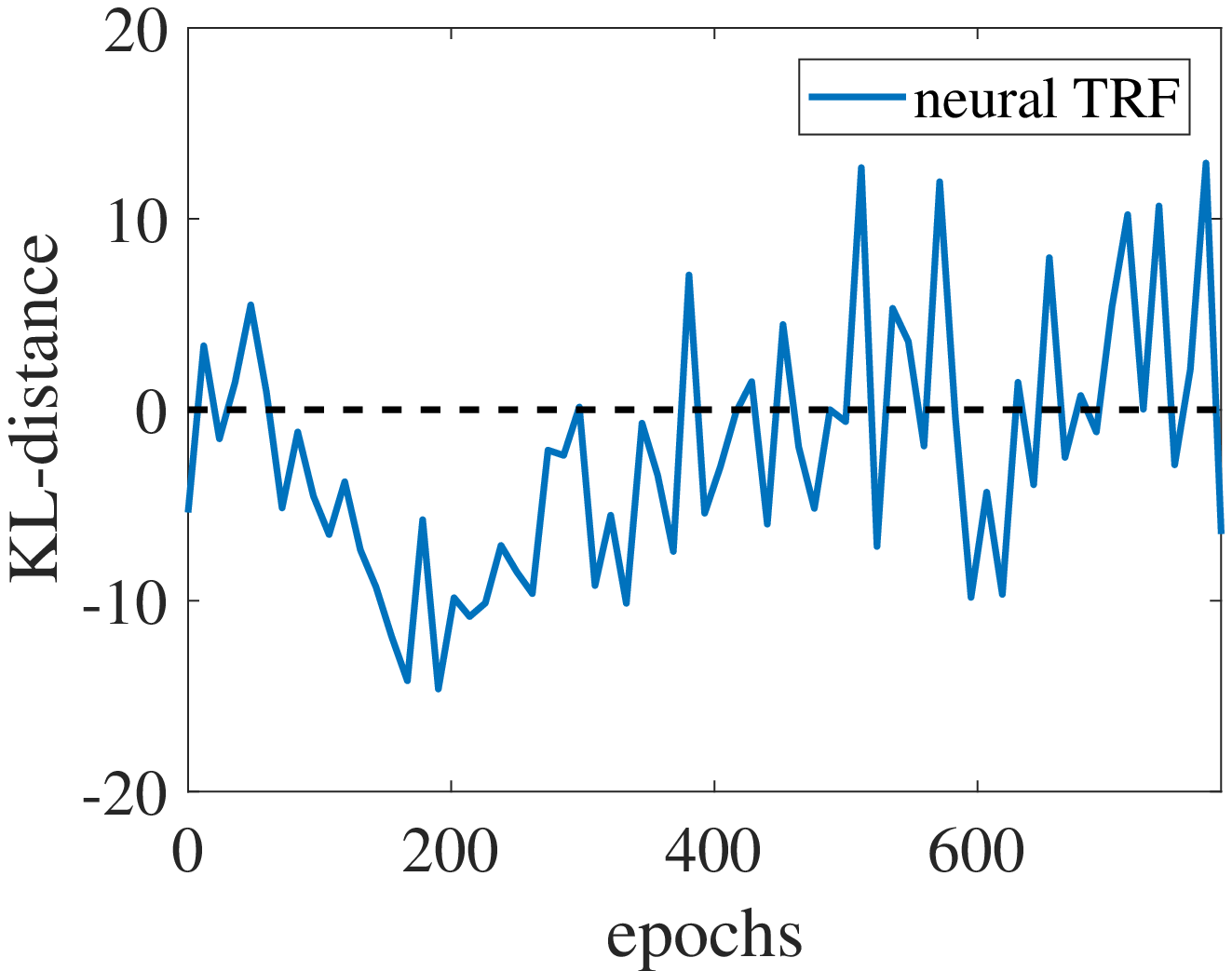}}
  \centerline{(b)}
\end{minipage}
\caption{(a) The negative log-likelihood on PTB test set and (b) the KL-divergence $KL(p||q)$, at each training epoch.}
\label{fig:plot}
\end{figure}

All the parameters of the CNN and LSTM are initialized randomly within an interval from -0.1 to 0.1, except for the word embedding of the CNN,
which is initialized by running the word2vec toolkit \cite{word2vec} on PTB training set and updated during training.
We stop the training once the smoothed log-likelihood (moving average of 1000 iterations) on the PTB development set does not increase significantly, resulting in 33,000 iterations (about 800 epochs).
The negative log-likelihood of TRF LMs on PTB test set and the KL-divergence between the model distribution $p(l,x^l;\theta,\zeta)$ and the auxiliary distribution $q(l,x^l;\mu)$ are shown in Fig. \ref{fig:plot}.

As the model parameters of neural TRF LMs are estimated stochastically, we cache 10 model parameters from the most recent 10 training epochs. After the training is stopped, we
calculate the PPLs over PTB test set and the LM scores of the 1000-best lists, using the cached 10 model parameters.
Then the resulting 10 PPLs are averaged as the final PPL, and the LM scores from the 10 models are averaged and used in rescoring, giving the final WER.
This is similar to model combination using sentence-level log-linear interpolation, which reduces the variance of stochastically estimated models.

The PPLs and WERs of various LMs are shown in Table \ref{tab:wer}, from which there are several comments.
First, as studied in \cite{Bin2017}, AugSA tends to underestimate the perplexity on test set.
The reported PPLs in Table \ref{tab:wer} should be a lower bound of the true PPLs of the TRF models.
Second, the neural TRF LM achieves the WER of 7.60, which outperforms the discrete TRF using features ``w+c+ws+cs+wsh+csh+tied" \cite{Bin2017} with relative reduction of 4.0\%.
Compared with ``LSTM-2x200'' with similar model size,
the neural TRF LM achieves a relative WER reduction of 4.5\%.
Compared with ``LSTM-2x650'',
the neural TRF LM achieves a slightly lower WER with only a fifth of parameters.
The large ``LSTM-2x1500'' performs slightly better than the neural TRF but with 16.5 times more parameters.
Third, we examine how neural TRF LMs and LSMT LMs are complimentary to each other. The probability of each sentences are log-linearly combined with equal interpolated weights of $0.5$.
The interpolated neural TRF and ``LSTM-2x1500'' further reduces the WER and achieves the lowest WER of 7.17.

Moreover, the inference with neural TRF LMs is much faster than with LSTM LMs.
The time cost of using ten neural TRFs to rescore the 1000-best list for each utterance is about 0.4 second.
Compared with LSTM LMs,
the inference of neural TRFs is about 16 times faster than ``LSTM-2x200'' and ``LSTM-2x650'',
and about 23 times faster than ``LSTM-2x1500''.

\section{Conclusion}
\label{sec:conclusion}

In this paper, we further study the TRF approach to language modeling and define neural TRF LMs that combine the advantages of both NNs and TRFs. The following contributions enable us to successfully train neural TRFs.
\begin{itemize}
  \item Although the nonlinear potential functions could be implemented by arbitrary NNs (FNNs or RNNs), the design of the deep CNN architecture in this paper is found to be important for efficient training.
  \item The proposed AugSA plus JSA training algorithm is crucial for the learning of neural TRFs to be feasible.
  \item Several additional techniques are found to be useful for training neural TRFs, including wider local jump in MCMC, Adam optimizer, and training set mini-batching.
\end{itemize}

It is worth pointing out that apart from the success in language modeling, the neural TRF models can also be applied to other sequential and trans-dimensional data modeling tasks in general, and also to discriminative modeling tasks, e.g. extending current ``CRFs+NNs" models.
For language modeling, integrating richer nonlinear and structured features is an important future direction.

\bibliographystyle{IEEEbib}
\bibliography{RF}

\end{document}

%% file: head.tex
\usepackage[nocompress]{cite}

\usepackage{times}
\usepackage{url}
\usepackage{latexsym}

\usepackage{amssymb} 
\usepackage{amsmath} 
\usepackage{amsthm}

\usepackage{array}
\usepackage{color}

\usepackage{algorithmicx}
\usepackage{algpseudocode}

\usepackage{graphicx}
\graphicspath{{fig/}}

\usepackage{multirow}
\newcommand{\equref}{\eqref}

\newcommand{\pdone}[1]{\frac{\partial}{\partial #1}}
\newcommand{\pd}[2]{\frac{\partial #1}{\partial #2}}